# Fully Automated and Standardized Segmentation of Adipose Tissue Compartments by Deep Learning in Three-dimensional Whole-body MRI of Epidemiological Cohort Studies

Thomas Küstner[1,2,3], PhD; Tobias Hepp[1,4], MD; Marc Fischer[1,2,5], M.Sc.; Martin Schwartz[2,5], M.Sc.; Andreas Fritsche[6,7,8], MD; Hans-Ulrich Häring[8], MD; Konstantin Nikolaou[1], MD; Fabian Bamberg[9], MD; Bin Yang[2], PhD; Fritz Schick[5,7,8], MD, PhD; Sergios Gatidis[1], MD; Jürgen Machann[5,7,8], PhD

[1] University Hospital Tübingen, Diagnostic and Interventional Radiology, Medical Image and Data Analysis, Tübingen, Germany

[2] University of Stuttgart, Department for Signal Processing and System Theory, Stuttgart, Germany

[3] School of Biomedical Engineering & Imaging Sciences, King's College London, St. Thomas' Hospital, London, UK

[4] Max-Planck Institute for Intelligent Systems, Empirical Inference Department, Tübingen, Germany

[5] University Hospital Tübingen, Diagnostic and Interventional Radiology, Section on Experimental Radiology, Tübingen, Germany

[6] Department of Internal Medicine IV, Eberhard Karls University, Tübingen, Germany

[7] Institute for Diabetes Research and Metabolic Diseases (IDM) of the Helmholtz Zentrum München at the University of Tübingen, Tübingen, Germany

[8] German Center for Diabetes Research (DZD)

[9] Department of Diagnostic and Interventional Radiology, Medical Center – University of Freiburg, Faculty of Medicine, University of Freiburg, Germany

Originating institution:

University Hospital of Tübingen

Hoppe-Seyler-Str. 3

72076 Tübingen, Germany

Corresponding author:

Thomas Küstner

Diagnostic and Interventional Radiology

Hoppe-Seyler-Str. 3

72076 Tübingen, Germany

e-mail: thomas.kuestner@med.uni-tuebingen.de

**Statement:**

**Fully automated and fast assessment of visceral and subcutaneous adipose tissue compartments in whole-body MRI is feasible by means of a deep learning network. A robust and generalizable architecture was investigated which enables objective segmentation and quick phenotypic profiling.**

**Key Points:**

- Objective, fast, and reliable assessment of adipose tissue compartments (subcutaneous and visceral fat) is feasible in MRI datasets for non-invasive and phenotypic adipose tissue profiling in less than 7 seconds.
- Robust and generalizable architecture to several imaging sequences, contrast agents, nine imaging sites, coil arrangement, patient positioning, three scanners, and two field strengths providing high Dice overlap (0.94) and classification sensitivity (96.6%), specificity (95.1%), precision (92.1%), and accuracy (98.4%).
- Integration into clinical workflow can support end-users, such as physicians and physiologists, to reduce time and effort for image analysis.

**Abbreviations**

CNN = convolutional neural network, DL = deep learning, DZD = German Center for Diabetes Research database, FSE = fast-spin echo, NAKO = German National Cohort database, 3D = three dimensional, TUEF = Tuebingen Family Study database, 2D = two-dimensional

## Abstract

**Purpose**: To enable fast and reliable assessment of subcutaneous and visceral adipose tissue compartments derived from whole-body MRI.

**Methods**: Quantification and localization of different adipose tissue compartments from whole-body MR images is of high interest to examine metabolic conditions. For correct identification and phenotyping of individuals at increased risk for metabolic diseases, a reliable automatic segmentation of adipose tissue into subcutaneous and visceral adipose tissue is required. In this work we propose a 3D convolutional neural network (DCNet) to provide a robust and objective segmentation. In this retrospective study, we collected 1000 cases (66±13 years; 523 women) from the Tuebingen Family Study and from the German Center for Diabetes research (TUEF/DZD), as well as 300 cases (53±11 years; 152 women) from the German National Cohort (NAKO) database for model training, validation, and testing with a transfer learning between the cohorts. These datasets had variable imaging sequences, imaging contrasts, receiver coil arrangements, scanners and imaging field strengths. The proposed DCNet was compared against a comparable 3D UNet segmentation in terms of sensitivity, specificity, precision, accuracy, and Dice overlap.

**Results**: Fast (5-7seconds) and reliable adipose tissue segmentation can be obtained with high Dice overlap (0.94), sensitivity (96.6%), specificity (95.1%), precision (92.1%) and accuracy (98.4%) from 3D whole-body MR datasets (field of view coverage 450x450x2000mm$^3$). Segmentation masks and adipose tissue profiles are automatically reported back to the referring physician.

**Conclusion**: Automatic adipose tissue segmentation is feasible in 3D whole-body MR data sets and is generalizable to different epidemiological cohort studies with the proposed DCNet.

## Introduction

MRI is a widely used imaging modality that enables highly resolved anatomical and functional depiction of organs, tissues, and disease processes. The acquisition of imaging data has become an integral part for reliable and fast disease assessment, staging, therapy, and treatment monitoring. Besides individual assessment of disease by MRI, large-scale cohort studies provide insight in factors influencing pathogenesis of various diseases (1, 2).

The increasing prevalence of type 2 diabetes mellitus demands further effort in large cohort MRI studies and fast automatic phenotyping to determine suitable biomarkers for risk assessment as well as to deliver personalized lifestyle intervention treatment for prevention. The adipose tissue (AT) distribution in the body has been shown to be a key indicator for the pathogenesis of insulin resistance and type 2 diabetes mellitus (3-5). Whole-body assessment of adipose tissue distribution by MRI has proven to be a promising non-invasive method for screening in large-scale cohort studies (6-8). One crucial step in the evaluation of medical image content is the recognition and segmentation of specific organs or tissues, i.e. performing a voxel-wise classification known as semantic segmentation. Several methods for automated segmentation of MR images have been proposed, including methods relying on explicit models (9), general correspondence (10), random forests (11), as well as deep learning (12) (DL) approaches including convolutional neural networks (CNNs) (13-15).. Most CNN-based segmentation architectures are derived from the UNet (15), due to its good generalizability and performance (16, 17). Landmark detection (18, 19) or anatomical object localization (20) have been demonstrated first and can be an important pre-processing step for segmentation. DL-based approaches on MRI data have been shown for segmentation of brain matters (21, 22), brain tumors (23, 24), liver (25), and cardiac ventricles (26).

Approaches for automatic adipose tissue segmentation have been proposed mainly by employing machine learning techniques (27). Suitable contrasts in MRI for differentiation of adipose tissue from all other tissue types is provided by the markedly T1-weighted imaging leading to hyperintense adipose tissue signal with respect to lean tissue. An alternative approach for fat-water imaging is phase sensitive to multi-echo gradient-echo (Dixon) imaging, exploiting the different chemical shifts from water to methylene and methyl signals from triglycerides (principle first described by Dixon (28)). The multi-echo acquisition enables to extract fat-water separated images and other confounding factors such as fat fraction or R2* maps (29-31). This information can be used to pre-process the data (eg bias field correction (32)) or can be incorporated into the segmentation. Adipose tissue segmentations include multi-atlas based machine learning for fat images of regional muscles (33), contour-based segmentations (34), and machine learning clustering methods (35). More recently two-dimensional (2D) DL networks for abdominal adipose tissue segmentation in Dixon MRI (36, 37) were proposed.

Adipose tissue segmentation on whole-body MRI enabled adipose tissue profiling (38) to investigate metabolic risks for cardiac disease (39), type 2 diabetes (40), or cancer (41).

However, the above works focused on either a single organ or body region, were performed on smaller scaled or single cohorts, or provided only a two-dimensional processing. In this work, we propose a semantic segmentation network for whole-body adipose tissue segmentation which operates on single- and multi-parametric 3D whole-body MRI from different multi-center epidemiological patient cohort studies. The aim of this work is to provide a segmentation network which is robust to datasets acquired in different imaging sites, with varying scanner types, field strengths, receiver coil setups, and for changing imaging sequences with varying imaging resolutions and different patient positioning.

This study includes four main objectives: *(a)* to develop a 3D DL architecture as a combination of a densely connected network with merge-and-run mappings for attention-based multi-resolution focusing in an encoder-decoder segmentation together with a relative positional encoding of input patches; *(b)* to determine its performance in terms of robustness, sensitivity, specificity, precision, accuracy and Dice overlap in comparison to a 3D UNet segmentation; *(c)* to assess the possibility of a transfer learning between different epidemiological cohorts for training database composition and impact on its performance; and *(d)* to integrate the model into clinical workflow with automated reporting of adipose tissue head-feet profiles to enable profiling and studying in an epidemiological setting.

## Material and Methods

A CNN is proposed for 3D semantic segmentation of whole-body adipose tissue into subcutaneous adipose tissue (SAT), abdominal visceral adipose tissue (VAT), lean tissue (LT) and background (BG). In order to cope with varying tissue appearances and distribution, the network utilizes a combination of architectural designs (14, 15, 42, 43) and proposes the inclusion of a positional encoding for an attention-based (44) multi-resolution learning (45) with enhanced receptive field of view (14, 15).

The network is trained and tested on whole-body MR data from different multi-center epidemiological patient databases with varying image contrast, imaging dimensionality, scanners, field strengths, patient positioning, and coil arrangement.

### *Epidemiological Patient Databases*

Database compositions, as well as scanner and acquisition parameters, are summarized in Table 1. Studies are approved by the ethics committees and individuals gave written consent. Inclusion and exclusion criteria and further study information are stated in (2, 6).

*TUEF and DZD Database*

The Tuebingen Family Study (TUEF) and the studies performed in the framework of the German Center for Diabetes Research (DZD) aim to measure fat distribution in Caucasian individuals at increased risk of type-2 diabetes due to being overweight (BMI > 27 kg/m$^2$), first degree relative of a patient type-2 diabetes, having impaired glucose tolerance, and/or gestational diabetes prior to and after lifestyle intervention (6). Data was acquired in a single-center (TUEF) and in a multi-center (DZD) study on 1.5T and 3T scanners with a whole-body T1-weighted multi breath-hold and multi-slice 2D fast spin echo (FSE) sequence in axial direction. Individuals were positioned in prone position (extended arms) with one rearrangement (first step was head first and second step was feet first). Data were recorded applying the whole-body coil of the MR unit. From the about 2000 scanned individuals in four sites, 500 (1.5T) and 500 (3T) were labeled and included in this study.

*NAKO database*

The German National Cohort (GNC; NAKO Gesundheitsstudie) aims to understand the natural history of a broad set of diseases and to potentially identify novel imaging biomarkers (2). The study randomly recruits individuals for an MRI screening in five imaging centers distributed throughout Germany. From the imaging protocol we evaluate the whole-body multi breath-hold 3D dual gradient echo chemical-shift (Dixon) sequence which was acquired on a 3T scanner in axial orientation. Individuals were positioned head first supine. In this setup, data recording was performed with local array receiver coils on the front and back of the individual. The scanner was equipped with an 18-channel body and a 32-channel spine coil of the manufacturer. So far about 15000 individuals had been examined at the time of this study, and 300 were labeled and included in this study.

**Ground Truth Labeling**

Training data are obtained from a semi-supervised labeling (34). An automatic fuzzy c-means clustering pre-segments the whole-body MR data based on intensity histograms after normalization and partial volume correction into background, lean, and adipose tissue. A subsequent snake algorithm divides the adipose tissue into SAT and VAT region. The obtained masks (BG, SAT, VAT, LT) were manually inspected and corrected (if necessary) by two trained experts with 9 (S.G.) and 17 (J.M.) years of experience in whole-body MRI on an in-house developed graphical user interface to ensure proper data curation.

**Proposed Architecture**

The CNN-based segmentation is inspired by the concepts developed in UNet (15) for pixel-wise localization, VNet (14) for volumetric medical image segmentation, ResNet (43) to cope with vanishing gradients and degradation problem, DenseNet (42) to enable deep supervision, and Merge-And-Run (46) to provide attention-based multi-resolution focusing. The proposed network architecture is depicted in Figure 1. It combines the aforementioned schemes which results in a densely connected CNN, denoted as DCNet in the following. Specific details on the DCNet architecture are described in Appendix E1 (supplement).

In total, the proposed DCNet consists of 154 layers which results in approximately 12 million trainable parameters for a single- and dual-channel 3D input. Each 3D volume is first normalized into a range of 0 to 1 and then cropped by a sliding window into overlapping 3D input patches of size 32 x 32 x 32 x $C$, whereas $C \in [1,2]$ is given by the multi-parametric input. For the T1-weighted FSE data $C = 1$, whereas for the multi-echo chemical-shift (Dixon) data $C = 2$ (fat and water image). An RMSprop with multi-class focal loss (47) and multi-metric loss (true-positive rate and Jaccard distance) is applied during training for a batch size of 48. Training was conducted over 100 epochs with an early stopping. The network was implemented in Tensorflow 1.14 and trained on a Nvidia Tesla V-100.

*Experiments*

The proposed method aims to reliably work in all cohorts and for changing input data (imaging sequences, contrasts, sites, scanners, field strengths, coil, and patient positioning). Therefore, robustness and generalizability of the proposed architecture towards changing training and test data are investigated. Robustness and reliability were investigated for all experiments by means of a 4-fold cross-validation for the proposed DCNet and compared against a 3D UNet segmentation (Appendix E2 [supplement]).

Different training, validation, and test set composition amongst and within epidemiological cohorts were created to infer if and what information can be exchanged. Furthermore, a transfer learning scheme was applied to the proposed DCNet to conduct the database domain change and compared against a multi-database training. The proposed network was trained on database A and tested on database B or C, in the following denoted by A → B|C. The following four scenarios were investigated: *(a)* A → A: Intra-database training and testing*, (b)* A → B: Inter-database testing to infer generalizability*, (c)* A+B → A|B: Transductive transfer learning with pre-training on database A (first 40 epochs) followed by fine-training on database B (remaining epochs) and testing on A or B, and *(d)* A&B → A|B multi-database learning for randomly shuffled samples from database A and B with testing on A or B. The latter case serves also as comparison if a guided learning (ie transfer learning) is superior than multi-database learning. Data compositions between the different epidemiological cohorts (TUEF/DZD and NAKO) as well as between the 1.5T and 3T scans of the TUEF/DZD database are investigated to infer influence on prediction for changing imaging sequences and contrast (T1-weighted FSE versus multi-echo chemical-shift Dixon), imaging sites and scanners, coil positioning, patient positioning (prone versus supine) and field strengths (1.5T versus 3T).

Training datasets for the training folds and for the smaller subsets are randomly selected from the labeled cohort to provide uniform distribution over phenotypic factors (sex, age, and body mass index). For all experiments, datasets were split into 70% for training, 10% for validation, and 20% for the test set. For each the 1.5T and 3T measurements (500 individuals for each) from TUEF/DZD, there were 350 training, 50 validation, and 100 test sets. For the NAKO set of 300 individuals, cases were split into 210 training, 30 validation, and 60 test sets. For each of the cross-validation runs, a different fixed training, validation, and test set was created.

### *Statistical Analysis*

All reported results represent (if not stated otherwise), the mean and two-sided standard deviation on all of the four cross-validation runs and test case. The performance was evaluated qualitatively and quantitatively in terms of specificity, sensitivity, precision, and accuracy (equations are found in Appendix E3) which are derived from the confusion matrix between the predicted segmentation $S$ and the labeled ground-truth $G$ of the four classes (SAT, VAT, LT, BG) on a voxel-by-voxel comparison in a one-against-rest approach (ie the target class is compared against the sum of all remaining classes). Segmentation overlaps are evaluated by means of Dice coefficient (equation found in Appendix E3) between the predicted segmentation $S$ and the labeled ground-truth $G$. Statistical significance for the performance metrics was determined with a paired Welch's t-test (significance level of α = .05) under the null hypothesis of equal means for unequal variances performed in Python (v3.6) with SciPy (v1.4.0).

### *Code Availability*

Code is made publicly available under https://github.com/lab-midas/med_segmentation.

## Results

Axial slices of the TUEF/DZD databases for individuals scanned on a 1.5T and 3T scanner are shown in Figure 2. The T1-weighted FSE images can be affected by spatial inhomogeneities of the B1-field and by inhomogeneous sensitivity characteristics of the receiver coil as pointed out by the arrows. The segmentation is robust towards these artifacts and in close agreement to the ground-truth. The same network architecture is also generalizable towards the NAKO database with individuals scanned on 3T scanners in multiple imaging sites as qualitatively depicted in Figure 3. Moreover, extension of the network input dimension from single channel (T1-weighted FSE) towards multi-channel (multi-echo chemical-shift; Dixon) preserves a stable segmentation indicating a robust architecture. Quantitative evaluation of these experiments is shown in Table 2. Low standard deviations for the proposed DCNet indicate good robustness for changing training data folds. On average for all segmented tissues and databases, a sensitivity of 96.6%, a specificity of 95.1%, a precision of 92.1% and an accuracy of 98.4% with a Dice of 0.94 was achieved. The proposed DCNet outperforms a standard 3D UNet with statistical significance by 30.2% for sensitivity, 40.1% ($P < .001$) for specificity, 151.6% ($P < .001$) for precision, and 46.9% ($P < .001$) for Dice score, except 16.0% ($P = .14$) for accuracy (Figure E1 [supplement]). Training of the proposed DCNet required approximately 25 hours, whereas only 5-7 seconds are needed for prediction after data set loading.

The influence of changing input data can be examined for changing scanners and imaging field strengths (1.5T and 3T) in Figure 4 and for different cohorts (TUEF/DZD vs NAKO; FSE vs Dixon; 2D vs 3D; prone vs supine; and for different scanners and receiver coil arrangements) in Figure 5. Intra-database training and testing (A→A) performs better than inter-database testing (A→B). Transfer learning (A+B→A|B) or mutli-database learning (A&B→A|B) overcome segmentation limitations of the inter-database experiments, like magnetic field inhomogeneities (Fig. 4) or different through-plane resolutions (Fig. 5: TUEF/DZD [10 mm] vs. NAKO [3 mm]) which did not exist in the respective training database. Quantitative scores for changing scanners and imaging field strengths are shown in Table 3 and for changing cohorts in Table 4 which substantiate these findings. The best performing experiment and hence best data utilization was obtained by a transfer learning (A+B) which was not better than multi-database learning (A&B) for sensitivity ($P = .94$ and .91), specificity ($P = .98$ and .90), precision ($P = .66$ and .95), accuracy ($P = .82$ and .67), and Dice overlap ($P = .86$ and .84) in Table 3 and Table 4, respectively.

The segmented images enable derivation of a head-feet adipose tissue profile which helps to quickly visualize the adipose tissue distribution as exemplary depicted in Figure E2 (supplement). These profiles together with the segmented masks are automatically reported back to the referring physician. The adipose tissue profiles also enable phenotypic characterization as shown for different test cases grouped by age and sex of the TUEF/DZD database in Figure 6 or grouped by body mass index and sex in Figure E3 (supplement). The respective adipose tissue profiles of the NAKO database are shown in Figure E4 and E5 (supplement). Lean tissue was in general elevated in men compared to women. Lean tissue and subcutaneous adipose tissue show similar percentage distribution in women in contrast to men.

## Discussion

In this work, we present a 3D adipose tissue segmentation network for whole-body MRI. The proposed network combines different architectural concepts and receives the positional input of the 3D patches to learn a multi-resolution attention-focusing. Reliable adipose tissue segmentation can be obtained with high Dice overlap (0.94), sensitivity (96.6%), specificity (95.1%), precision (92.1%), and accuracy (98.4%) which is also significantly better than a comparable UNet segmentation. A fast, whole-body AT segmentation can be conducted within 5-7 seconds which enables the integration into clinical workflow. The returned segmentation masks and adipose tissue profiles enable a quick phenotypic assessment. We investigated its performance for varying training databases (TUEF/DZD and NAKO) with changing scanners and imaging field strengths (1.5T and 3T), receiver coil setups, imaging sequences (multi-slice 2D T1-weighted FSE and 3D multi-echo chemical-shift Dixon), and patient positioning (prone and supine). Impact of changing training and test data as well as information sharing between and within epidemiological cohort studies by means of transfer learning is investigated. The aim was to provide a robust architecture which generalizes well.

The proposed network deals well with changing input data such as single- and multi-contrast images, varying resolutions, scanners, and imaging field strengths. The network performance can vary in testing if the image content differs strongly from the trained cases (eg TUEF/DZD → NAKO) because the visual appearance of adipose tissue in FSE and DIXON images is diverse. This obstacle can be mitigated by transfer or joint learning on the data. We observed a minor improvement of transfer learning (A+B) over a multi-database learning (A&B) (ie guiding the network training with apriori knowledge improves its performance). Moreover, utilization of complementary information can help to mitigate segmentation errors (eg magnetic field inhomogeneities in 3T cases as observed in Figure 4). If the network is trained on artifact-affected images, segmentation on good cases becomes more challenging (eg, 3T → 1.5T). These results suggest that a careful selection and training database composition is important for reliable and robust segmentation.

Segmentation of background performs equally well in all experiments with high sensitivity, specificity, and precision. Visceral adipose tissue is more challenging to segment than subcutaneous adipose tissue, because of its patient-dependent appearance and irregular structure. Nevertheless, high metrics were achieved as well. Performance of 1.5T cases in TUEF/DZD was slightly better, because of less likely disturbance by magnetic field inhomogeneities.

The proposed architecture outperforms a standard UNet due to its multi-resolution sharing and attention focusing. For UNet, we observed repetitive patterns of misclassification which might indicate insufficient network depth for given input dimensionality. The applied 3D patching in DCNet provides invariance to epidemiological parameters such as height and weight and no additional pre-processing like image alignment is required. Furthermore, invariance to patient positioning (prone versus supine) was observed. Overall good generalization to different scan conditions (imaging sequences, scanners, coil, and patient positioning) and anthropometric features were obtained. Results from cross-validation runs with different training and test cases showed small standard deviations indicating reliable and robust training. Based on these experiences, we anticipate good reproducibility for repeated predictions and measurements.

Our study has limitations. The obtained results are specific for the imaging setup and MR sequence design of the study at hand. The segmented tissues in our study are rather large structures and may thus be easier to segment compared to more complicated anatomic structures. Thus, generalizability to other segmentation tasks and methods will be investigated in future studies. The network is currently not trained to distinguish bone marrow, which leads to misclassifications. However, for the underlying task of adipose tissue profiling, these misclassifications only attribute to a small and almost constant amount amongst individuals. In future studies, we plan to extend the labeling and classification with a bone marrow class.

In conclusion, automatic 3D AT segmentation and standardized topography mapping is feasible in whole-body MRI data with the proposed DCNet. The proposed architecture utilizes merge-and-run mapping blocks, dense connections, and patch-input encoding to providing a multi-resolution attention focusing. The architecture is robust and generalizable to different imaging sequences, contrasts, sites, scanners, field strengths and examination setups (coil arrangements and patient positioning). Segmented adipose tissue compartments (subcutaneous and visceral fat) and head-feet profiles are automatically generated and provide feedback to the referring physician.

## Acknowledgements

The work was supported in part by a grant (01GI0925) from the German Federal Ministry of Education and Research (BMBF) to the German Center for Diabetes Research (DZD e.V.) and by the Deutsche Forschungsgemeinschaft (DFG, German Research Foundation) – 428224476 / SPP 2177.

This project was conducted with data from the German National Cohort (GNC) (www.nako.de). The GNC is funded by the Federal Ministry of Education and Research (BMBF) [project funding reference numbers: 01ER1301A/B/C and 01ER1511D], federal states and the Helmholtz Association with additional financial support by the participating universities and the institutes of the Leibniz Association. We thank all participants who took part in the GNC study and the staff in this research program.

# References

1.	Machann J, Thamer C, Stefan N, et al. Follow-up Whole-Body Assessment of Adipose Tissue Compartments during a Lifestyle Intervention in a Large Cohort at Increased Risk for Type 2 Diabetes. Radiology. 2010;257(2):353-63.
2.	Bamberg F, Kauczor HU, Weckbach S, et al. Whole-Body MR Imaging in the German National Cohort: Rationale, Design, and Technical Background. Radiology. 2015;277(1):206-20.
3.	Kissebah AH, Vydelingum N, Murray R, Evans DJ, Kalkhoff RK, Adams PW. Relation of body fat distribution to metabolic complications of obesity. The Journal of Clinical Endocrinology & Metabolism. 1982;54(2):254-60.
4.	Krotkiewski M, Björntorp P, Sjöström L, Smith U. Impact of obesity on metabolism in men and women. Importance of regional adipose tissue distribution. The Journal of clinical investigation. 1983;72(3):1150-62.
5.	Ohlson L-O, Larsson B, Svärdsudd K, et al. The influence of body fat distribution on the incidence of diabetes mellitus: 13.5 years of follow-up of the participants in the study of men born in 1913. Diabetes. 1985;34(10):1055-8.
6.	Machann J, Thamer C, Schnoedt B, et al. Standardized assessment of whole body adipose tissue topography by MRI. Journal of Magnetic Resonance Imaging: An Official Journal of the International Society for Magnetic Resonance in Medicine. 2005;21(4):455-62.
7.	Linge J, Borga M, West J, et al. Body Composition Profiling in the UK Biobank Imaging Study. Obesity. 2018;26(11):1785-95.
8.	Linge J, Whitcher B, Borga M, Dahlqvist Leinhard O. Sub-phenotyping metabolic disorders using body composition: an individualized, nonparametric approach utilizing large data sets. Obesity. 2019;27(7):1190-9.
9.	Heimann T, Meinzer H-P. Statistical shape models for 3D medical image segmentation: a review. Medical image analysis. 2009;13(4):543-63.
10.	Iglesias JE, Sabuncu MR. Multi-atlas segmentation of biomedical images: a survey. Medical image analysis. 2015;24(1):205-19.
11.	Criminisi A, Shotton J, Konukoglu E, others. Decision forests: A unified framework for classification, regression, density estimation, manifold learning and semi-supervised learning. Foundations and Trends in Computer Graphics and Vision. 2012;7(2--3):81-227.
12.	Shen D, Wu G, Suk H-I. Deep learning in medical image analysis. Annual review of biomedical engineering. 2017;19:221-48.
13.	Litjens G, Kooi T, Bejnordi BE, et al. A survey on deep learning in medical image analysis. Medical image analysis. 2017;42:60-88.
14.	Milletari F, Navab N, Ahmadi S-A. V-Net: Fully Convolutional Neural Networks for Volumetric Medical Image Segmentation. ArXiv e-prints. 2016.
15.	Ronneberger O, Fischer P, Brox T. U-net: Convolutional networks for biomedical image segmentation. International Conference on Medical Image Computing and Computer-Assisted Intervention. 2015:234-41.
16.	Çiçek Ö, Abdulkadir A, Lienkamp SS, Brox T, Ronneberger O. 3D U-Net: learning dense volumetric segmentation from sparse annotation. International conference on medical image computing and computer-assisted intervention. 2016:424-32.
17.	Ibtehaz N, Rahman MS. MultiResUNet: Rethinking the U-Net Architecture for Multimodal Biomedical Image Segmentation. arXiv preprint arXiv:190204049. 2019.
18.	Yang D, Zhang S, Yan Z, Tan C, Li K, Metaxas D. Automated anatomical landmark detection ondistal femur surface using convolutional neural network. 2015 IEEE 12th international symposium on biomedical imaging (ISBI). 2015:17-21.
19.	Payer C, Štern D, Bischof H, Urschler M. Regressing heatmaps for multiple landmark localization using CNNs. International Conference on Medical Image Computing and Computer-Assisted Intervention. 2016:230-8.

20. de Vos BD, Wolterink JM, de Jong PA, Viergever MA, Išgum I. 2D image classification for 3D anatomy localization: employing deep convolutional neural networks. Medical Imaging 2016: Image Processing. 2016;9784:97841Y.
21. Zhang W, Li R, Deng H, et al. Deep convolutional neural networks for multi-modality isointense infant brain image segmentation. NeuroImage. 2015;108:214-24.
22. Moeskops P, Viergever MA, Mendrik AM, de Vries LS, Benders MJ, Išgum I. Automatic segmentation of MR brain images with a convolutional neural network. IEEE transactions on medical imaging. 2016;35(5):1252-61.
23. Pereira S, Pinto A, Alves V, Silva CA. Brain tumor segmentation using convolutional neural networks in MRI images. IEEE transactions on medical imaging. 2016;35(5):1240-51.
24. Havaei M, Davy A, Warde-Farley D, et al. Brain tumor segmentation with deep neural networks. Medical image analysis. 2017;35:18-31.
25. Qin W, Wu J, Han F, et al. Superpixel-based and boundary-sensitive convolutional neural network for automated liver segmentation. Physics in Medicine & Biology. 2018;63(9):095017.
26. Bai W, Sinclair M, Tarroni G, et al. Automated cardiovascular magnetic resonance image analysis with fully convolutional networks. Journal of Cardiovascular Magnetic Resonance. 2018;20(1):65.
27. Borga M. MRI adipose tissue and muscle composition analysis-a review of automation techniques. The British journal of radiology. 2018;91(1089):20180252.
28. Dixon WT. Simple proton spectroscopic imaging. Radiology. 1984;153(1):189-94.
29. Yu H, Shimakawa A, McKenzie CA, Brodsky E, Brittain JH, Reeder SB. Multiecho water-fat separation and simultaneous $R2^*$ estimation with multifrequency fat spectrum modeling. Magnetic Resonance in Medicine. 2008;60(5):1122-34.
30. Reeder SB, Hu HH, Sirlin CB. Proton density fat-fraction: a standardized MR-based biomarker of tissue fat concentration. Journal of magnetic resonance imaging : JMRI. 2012;36(5):1011-4.
31. Hu HH, Chen J, Shen W. Segmentation and quantification of adipose tissue by magnetic resonance imaging. Magma. 2016;29(2):259-76.
32. Vovk U, Pernus F, Likar B. A review of methods for correction of intensity inhomogeneity in MRI. IEEE Trans Med Imaging. 2007;26(3):405-21.
33. Karlsson A, Rosander J, Romu T, et al. Automatic and quantitative assessment of regional muscle volume by multi-atlas segmentation using whole-body water-fat MRI. Journal of magnetic resonance imaging : JMRI. 2015;41(6):1558-69.
34. Würslin C, Machann J, Rempp H, Claussen C, Yang B, Schick F. Topography mapping of whole body adipose tissue using A fully automated and standardized procedure. Journal of Magnetic Resonance Imaging. 2010;31(2):430-9.
35. Addeman BT, Kutty S, Perkins TG, et al. Validation of volumetric and single-slice MRI adipose analysis using a novel fully automated segmentation method. Journal of Magnetic Resonance Imaging. 2015;41(1):233-41.
36. Estrada S, Lu R, Conjeti S, et al. FatSegNet: A fully automated deep learning pipeline for adipose tissue segmentation on abdominal dixon MRI. Magnetic Resonance in Medicine;n/a(n/a).
37. Langner T, Hedstrom A, Morwald K, et al. Fully convolutional networks for automated segmentation of abdominal adipose tissue depots in multicenter water-fat MRI. Magn Reson Med. 2019;81(4):2736-45.
38. Borga M, West J, Bell JD, et al. Advanced body composition assessment: from body mass index to body composition profiling. Journal of Investigative Medicine. 2018;66(5):1.
39. Artham SM, Lavie CJ, Patel HM, Ventura HO. Impact of obesity on the risk of heart failure and its prognosis. J Cardiometab Syndr. 2008;3(3):155-61.
40. Kurioka S, Murakami Y, Nishiki M, Sohmiya M, Koshimura K, Kato Y. Relationship between visceral fat accumulation and anti-lipolytic action of insulin in patients with type 2 diabetes mellitus. Endocr J. 2002;49(4):459-64.
41. Doyle SL, Donohoe CL, Lysaght J, Reynolds JV. Visceral obesity, metabolic syndrome, insulin resistance and cancer. Proc Nutr Soc. 2012;71(1):181-9.

42. Huang G, Liu Z, Weinberger KQ, van der Maaten L. Densely connected convolutional networks. arXiv preprint arXiv:160806993. 2016.
43. He K, Zhang X, Ren S, Sun J. Deep residual learning for image recognition. Proceedings of the IEEE conference on computer vision and pattern recognition. 2016:770-8.
44. Vaswani A, Shazeer N, Parmar N, et al. Attention is all you need. Advances in neural information processing systems. 2017:5998-6008.
45. Yu F, Koltun V. Multi-scale context aggregation by dilated convolutions. arXiv preprint arXiv:151107122. 2015.
46. Zhao L, Wang J, Li X, Tu Z, Zeng W. Deep convolutional neural networks with merge-and-run mappings. arXiv preprint arXiv:161107718. 2016.
47. Lin T, Goyal P, Girshick R, He K, Dollár P. Focal Loss for Dense Object Detection. 2017 IEEE International Conference on Computer Vision (ICCV). 2017:2999-3007.

**Tables**

Table 1: TUEF/DZD and NAKO Database Information

| Characteristic | TUEF/DZD | NAKO |
|---|---|---|
| All individuals (*n*) | 1000 | 300 |
|   Age (y) | 56 ± 13 (18–76) | 53 ± 11 (22–72) |
|   BMI (kg/m$^2$) | 31.7 ± 6.8 (20.2–42.9) | 29.1 ± 6.2 (18.0–52.6) |
| Women (*n*) | 523 | 152 |
|   Age (y) | 53 ± 13 (18–76) | 54 ± 10 (22–72) |
|   BMI (kg/m$^2$) | 33.0 ± 7.3 (20.2–42.9) | 28.8 ± 6.9 (18.6–52.6) |
| Men (*n*) | 477 | 148 |
|   Age (y) | 51 ± 11 (19–75) | 53 ± 11 (22–72) |
|   BMI (kg/m$^2$) | 29.5 ± 5.0 (20.4–42.2) | 29.3 ± 5.6 (18.0–49.5) |
| Sequence | Fast Spin Echo | Dixon |
| FOV (mm3) | 450 x 450 x 2000 | 448 x 364 x 948 |
| Resolution (mm) | 2 x 2 x 10 | 1.4 x 1.4 x 3 |
| 3D Image Size | 225 x 225 x 200 | 320 x 260 x 316 |
| TE (ms) | 12 | 1.23 and 2.46 |
| TR (ms) | 490 | 4.36 |
| Flip angle | 170° | 9° |
| Bandwidth (Hz/px) | 130 | 680 |
| Patient positioning | Prone | Supine |
| 1.5T (*n*)[*] | 500 | 0 |
|   Magnetom Sonata | 218 | 0 |
|   Magnetom Avanto | 282 | 0 |
| 3T (*n*)[*] | 500 | 300 |
|   Magnetom Prisma | 398 | 0 |
|   Magnetom Vida | 102 | 0 |
|   Magnetom Skyra | 0 | 300 |

Note.— The average ± standard deviation (range) is shown for age and body mass index (BMI). 3D = three-dimensional, TE = echo time, TR = repetition time

[*]Scanners are from Siemens Healthcare, Erlangen, Germany.

Küstner et al. *Radiology Artificial Intelligence* 2020Table 2. Quantitative Evaluation Metrics for TUEF/DZD and NAKO Test Cases Over All Cross-validation Runs

| | Database | Sensitivity | | | | Specificity | | | |
|---|---|---|---|---|---|---|---|---|---|
| | | BG | LT | VAT | SAT | BG | LT | VAT | SAT |
| Proposed DCNet | TUEF/DZD (1.5T) | 99.7% ± 0.2% | 98.4% ± 0.3% | 91.9% ± 0.2% | 98.4% ± 0.1% | 99.9% ± 0.02% | 92.4% ± 0.4% | 90.6% ± 1.1% | 97.3% ± 0.8% |
| | TUEF/DZD (3T) | 99.2% ± 0.1% | 97.0% ± 0.5% | 92.3% ± 0.2% | 97.1% ± 0.1% | 99.9% ± 0.01% | 92.6% ± 0.3% | 90.4% ± 0.9% | 97.3% ± 0.7% |
| | NAKO (3T) | 99.7% ± 0.2% | 97.7% ± 0.3% | 90.2% ± 0.2% | 97.7% ± 0.1% | 99.9% ± 0.01% | 91.5% ± 0.5% | 91.5% ± 0.7% | 97.7% ± 0.6% |
| UNet | TUEF/DZD (1.5T) | 98.9% ± 1.2% | 74.1% ± 6.7% | 63.2% ± 12.5% | 76.8% ± 9.9% | 86.3% ± 6.4% | 63.5% ± 12.2% | 59.5% ± 9.6% | 67.5% ± 14.5% |
| | TUEF/DZD (3T) | 91.7% ± 7.2% | 72.5% ± 19.2% | 58.4% ± 24.5% | 72.7% ± 14.3% | 85.1% ± 6.9% | 64.7% ± 23.1% | 61.2% ± 28.4% | 67.3% ± 27.1% |
| | NAKO (3T) | 80.4% ± 12.3% | 73.2% ± 21.3% | 59.3% ± 32.1% | 69.6% ± 22.4% | 86.3% ± 9.9% | 57.2% ± 21.1% | 55.2% ± 24.1% | 61.2% ± 19.8% |

| | | Precision | | | | Dice | | | |
|---|---|---|---|---|---|---|---|---|---|
| | | BG | LT | VAT | SAT | BG | LT | VAT | SAT |
| Proposed DCNet | TUEF/DZD (1.5T) | 99.0% ± 0.1% | 91.3% ± 0.8% | 91.5% ± 0.7% | 90.3% ± 0.6% | 0.99 ± 0.01 | 0.94 ± 0.02 | 0.92 ± 0.02 | 0.98 ± 0.02 |
| | TUEF/DZD (3T) | 95.4% ± 0.4% | 89.4% ± 0.7% | 90.5% ± 1.1% | 90.3% ± 0.6% | 0.99 ± 0.01 | 0.90 ± 0.04 | 0.86 ± 0.06 | 0.94 ± 0.03 |
| | NAKO (3T) | 97.4% ± 0.2% | 90.4% ± 0.4% | 89.4% ± 1.2% | 90.3% ± 0.8% | 0.99 ± 0.01 | 0.91 ± 0.03 | 0.89 ± 0.05 | 0.95 ± 0.05 |
| UNet | TUEF/DZD (1.5T) | 55.2% ± 24.8% | 32.2% ± 29.7% | 34.1% ± 28.8% | 33.1% ± 26.8% | 0.91 ± 0.06 | 0.73 ± 0.08 | 0.51 ± 0.14 | 0.65 ± 0.22 |
| | TUEF/DZD (3T) | 52.5% ± 26.2% | 30.7% ± 27.1% | 28.2% ± 29.1% | 31.5% ± 28.6% | 0.78 ± 0.10 | 0.64 ± 0.11 | 0.44 ± 0.21 | 0.58 ± 0.18 |
| | NAKO (3T) | 49.4% ± 25.4% | 33.9% ± 28.2% | 31.2% ± 25.2% | 27.6% ± 31.3% | 0.81 ± 0.09 | 0.61 ± 0.13 | 0.43 ± 0.17 | 0.55 ± 0.16 |

| Accuracy | TUEF/DZD (1.5T) | TUEF/DZD (3T) | NAKO (3T) |
|---|---|---|---|
| Proposed DCNet | 99.0% ± 0.4% | 98.0% ± 0.3% | 98.3% ± 0.2% |
| UNet | 91.1% + 8.9% | 88.1% ± 7.4% | 75.2% ± 19.2% |

Note.— Quantitative evaluation metrics for TUEF/DZD (scanned on 1.5T and 3T scanners) and NAKO (scanned on 3T scanners) test cases as average and standard deviation over all cross-validation runs. Networks were trained on the individual databases (TUEF/DZD or NAKO). Sensitivity, specificity, precision, accuracy and Dice overlap measure the performance between predicted segmentation and labelled ground-truth in all segmented classes: BG = background, LT = lean tissue, SAT = subcutaneous adipose tissue, VAT = visceral adipose tissue. The DCNet outperformed the UNet for all metrics.

This manuscript has been accepted for publication in *Radiology: Artificial Intelligence* (https://pubs.rsna.org/journal/ai), which is published by the Radiological Society of North America (RSNA).

Küstner et al. *Radiology Artificial Intelligence* 2020

Table 3. Quantitative Evaluation Metrics of Proposed DCNet for Different Training and Testing Pairings between 1.5T and 3T Test Cases in TUEF/DZD Database

| Database | Sensitivity | | | | Specificity | | | |
|---|---|---|---|---|---|---|---|---|
| | BG | LT | VAT | SAT | BG | LT | VAT | SAT |
| 1.5T → 1.5T<br>3T → 3T | 99.6%<br>± 0.1% | 98.2%<br>± 0.2%* | 93.9%<br>± 0.2%* | 98.4%<br>± 0.1%* | 99.9%<br>± 0.02% | 92.4%<br>± 0.4% | 90.6%<br>± 1.1% | 97.3%<br>± 0.8% |
| 1.5T → 3T<br>3T → 1.5T | 99.1%<br>± 0.7% | 95.4%<br>± 0.9% | 86.3%<br>± 1.2% | 94.0%<br>± 1.1% | 99.9%<br>± 0.9% | 92.5%<br>± 1.3% | 90.4%<br>± 1.9% | 97.3%<br>± 0.9% |
| 1.5T + 3T → 1.5T<br>1.5T + 3T → 3T | 99.9%<br>± 0.2%* | 97.8%<br>± 0.4% | 92.2%<br>± 0.3% | 97.9%<br>± 0.1% | 99.9%<br>± 0.01%* | 92.6%<br>± 0.4%* | 90.6%<br>± 0.6%* | 97.5%<br>± 0.5%* |
| 1.5T & 3T → 1.5T<br>1.5T & 3T → 3T | 99.5%<br>± 0.2% | 97.7%<br>± 0.3% | 92.1%<br>± 0.2% | 97.8%<br>± 0.1% | 99.9%<br>± 0.01% | 92.5%<br>± 0.5% | 90.5%<br>± 0.7% | 97.4%<br>± 0.6% |
| | Precision | | | | Dice | | | |
| | BG | LT | VAT | SAT | BG | LT | VAT | SAT |
| 1.5T → 1.5T<br>3T → 3T | 99.0%<br>± 0.1%* | 91.5%<br>± 0.8% | 91.2%<br>± 0.7% | 92.3%<br>± 0.6% | 0.99<br>± 0.01* | 0.92<br>± 0.04 | 0.89<br>± 0.03 | 0.96<br>± 0.03* |
| 1.5T → 3T<br>3T → 1.5T | 93.4%<br>± 1.4% | 87.2%<br>± 1.4% | 85.1%<br>± 1.8% | 89.4%<br>± 1.2% | 0.99<br>± 0.04 | 0.89<br>± 0.04 | 0.79<br>± 0.12 | 0.89<br>± 0.07 |
| 1.5T + 3T → 1.5T<br>1.5T + 3T → 3T | 97.4%<br>± 0.2% | 91.8%<br>± 0.4%* | 91.4%<br>± 0.6%* | 92.3%<br>± 0.5%* | 0.99<br>± 0.02 | 0.92<br>± 0.03* | 0.90<br>± 0.04* | 0.95<br>± 0.03 |
| 1.5T & 3T → 1.5T<br>1.5T & 3T → 3T | 97.2%<br>±0.3% | 90.4%<br>±0.5% | 91.0%<br>±0.5% | 90.3%<br>±0.7% | 0.99<br>± 0.03 | 0.91<br>± 0.02 | 0.90<br>± 0.05 | 0.94<br>± 0.02 |
| | 1.5T → 1.5T<br>3T → 3T | | 1.5T → 3T<br>3T → 1.5T | | 1.5T + 3T → 1.5T<br>1.5T + 3T → 3T | | 1.5T & 3T → 1.5T<br>1.5T & 3T → 3T | |
| Accuracy | 98.8% ± 0.4% | | 97.6% ± 0.6% | | 99.1% ± 0.2%* | | 98.5% + 0.3% | |

Note.— Quantitative evaluation metrics of proposed DCNet for TUEF/DZD individuals (scanned on 1.5T and 3T scanners) for training on database A and testing on database B or C, denoted as A → B|C. Average and standard deviation of test cases selected from test sets of database B and C and over all cross-validation runs are shown. Intra- (A→A) and inter-data (A → B) training and testing as well as transfer learning (A+B → A|B) or multi-database learning (A&B → A|B) were investigated. Sensitivity, specificity, precision, Dice overlap, and accuracy measure the performance between predicted segmentation and labelled ground-truth in all segmented classes: BG = background, LT = lean tissue, SAT = subcutaneous adipose tissue, VAT = visceral adipose tissue.

*Indicates the best performing experiment.

Küstner et al. *Radiology Artificial Intelligence* 2020

Table 4. Quantitative Evaluation Metrics of Proposed DCNet for Different Training and Testing Pairings between TUEF/DZD (T/D) and NAKO (N) Databases

| Database | Sensitivity | | | | Specificity | | | |
|---|---|---|---|---|---|---|---|---|
| | BG | LT | VAT | SAT | BG | LT | VAT | SAT |
| T/D → T/D<br>N → N | 99.6% ± 0.1% | 97.9% ± 0.3% | 92.9% ± 0.2%* | 98.4% ± 0.2%* | 99.9% ± 0.02% | 92.5% ± 0.4% | 91.6% ± 1.0%* | 97.8% ± 0.7%* |
| T/D → N<br>N → T/D | 99.0% ± 1.1% | 94.4% ± 1.9% | 85.3% ± 2.2% | 89.0% ± 4.1% | 98.9% ± 0.9% | 82.6% ± 5.3% | 80.4% ± 6.9% | 87.3% ± 9.2% |
| T/D + N → T/D<br>T/D + N → N | 99.7% ± 0.2%* | 97.9% ± 0.3%* | 92.2% ± 0.2% | 97.7% ± 0.1% | 99.9% ± 0.01%* | 92.9% ± 0.4%* | 91.5% ± 0.7% | 97.7% ± 0.6% |
| T/D & N → T/D<br>T/D & N → N | 99.6% ± 0.3% | 97.7% ± 0.4% | 91.2% ± 0.3% | 97.8% ± 0.1% | 99.9% ± 0.04% | 92.0% ± 0.3% | 91.0% ± 0.8% | 97.6% ± 0.9% |

| Database | Precision | | | | Dice | | | |
|---|---|---|---|---|---|---|---|---|
| | BG | LT | VAT | SAT | BG | LT | VAT | SAT |
| T/D → T/D<br>N → N | 98.0% ± 0.3%* | 90.4% ± 0.6% | 90.3% ± 0.5% | 90.5% ± 0.8%* | 0.99 ± 0.02 | 0.92 ± 0.03 | 0.89 ± 0.04 | 0.96 ± 0.02 |
| T/D → N<br>N → T/D | 88.4% ± 5.4% | 81.2% ± 7.4% | 75.1% ± 9.8% | 85.4% ± 6.2% | 0.98 ± 0.02 | 0.88 ± 0.06 | 0.82 ± 0.07 | 0.89 ± 0.05 |
| T/D + N → T/D<br>T/D + N → N | 97.4% ± 0.2% | 90.6% ± 0.4%* | 90.4% ± 0.6%* | 90.4% ± 0.7% | 0.99 ± 0.01* | 0.93 ± 0.04* | 0.92 ± 0.02* | 0.97 ± 0.03* |
| T/D & N → T/D<br>T/D & N → N | 97.3% ±0.5% | 90.4% ± 0.7% | 90.2% ± 0.4% | 90.3% ± 0.2% | 0.99 ± 0.02 | 0.93 ± 0.05 | 0.91 ± 0.02 | 0.96 ± 0.03 |

| | T/D → T/D<br>N → N | T/D → N<br>N → T/D | T/D + N → T/D<br>T/D + N → N | T/D & N → T/D<br>T/D & N → N |
|---|---|---|---|---|
| Accuracy | 98.8% ± 0.6% | 87.6% ± 2.6% | 98.8% ± 0.2%* | 98.4% + 0.4% |

Note.— Quantitative evaluation metrics of proposed DCNet for TUEF/DZD (T/D, two-dimensional T1-weighted fast spin echo in prone position) and NAKO (N, three-dimensional multi-echo chemical-shift Dixon in supine position) for training on database A and tested on database B or C, denoted as A → B|C. Average and standard deviation of test cases selected from test sets of database B and C and over all cross-validation runs are shown. Intra- (A→A) and inter-data (A → B) training and testing as well as transfer learning (A+B → A|B) or multi-database learning (A and B → A|B) are investigated. Sensitivity, specificity, precision, Dice overlap, and accuracy measure the performance between predicted segmentation and labelled ground-truth in all segmented classes: BG = background, LT = lean tissue, SAT = subcutaneous adipose tissue, VAT = visceral adipose tissue.

*Indicates the best performing experiment.

**Figure Legends**

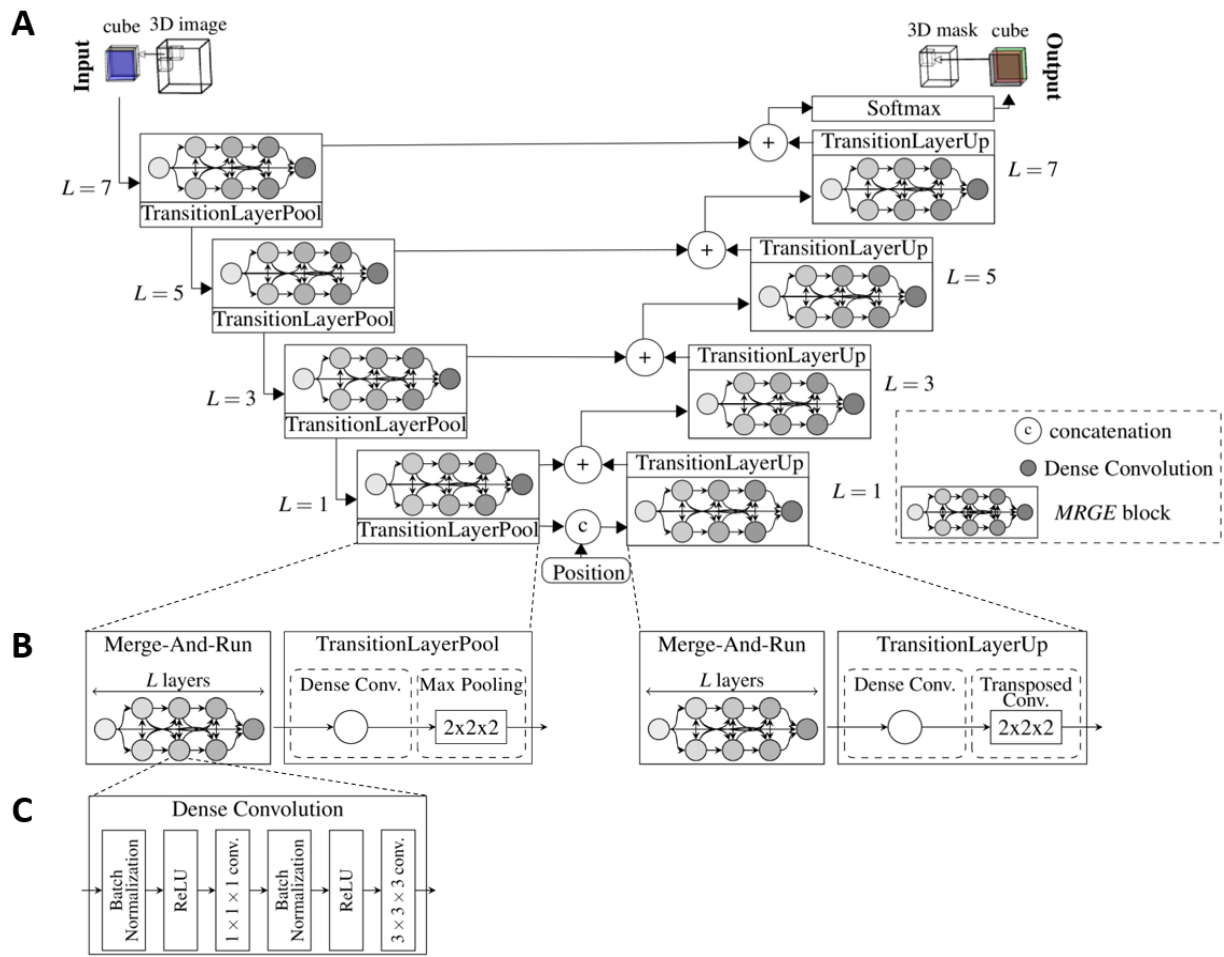

Figure 1. Proposed three-dimensional DCNet segmentation network. **(a)** General encoder-decoder structure which consists of Merge-And-Run (MRGE) blocks and intermittent transition layers for downsampling (encoder path) and upsampling (decoder path) of feature maps to provide multi-resolution segmentation. MRGE blocks are built up with dense connections for deep supervision and *L* layers of Dense Convolution nodes. **(b)** Composition of MRGE block with dense merge-and-run connections and transition layers in encoder path (max-pooling) and in decoder path (transposed convolution). **(c)** Each dense convolution node is a series of batch normalization, rectified linear unit (ReLU) activation, 1 x 1 x 1 convolution, batch normalization, ReLU activation and 3 x 3 x 3 convolution.

Küstner et al. *Radiology Artificial Intelligence* 2020

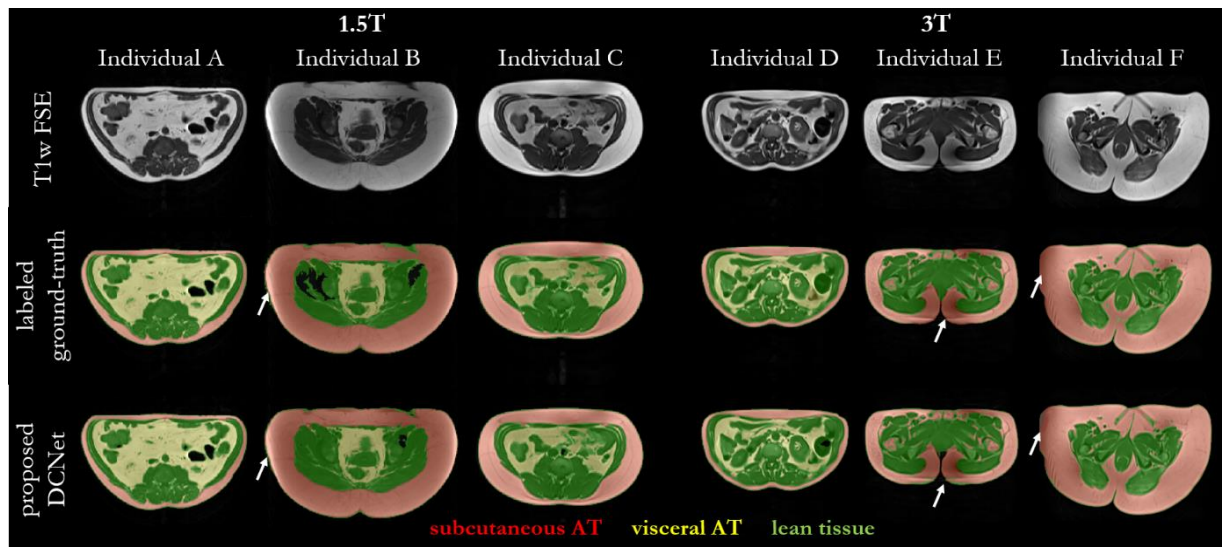

Figure 2. Adipose tissue (AT) segmentation into subcutaneous AT (red), visceral AT (yellow), lean tissue (green), and background (black) in three individuals scanned on a 1.5T scanner and in three individuals scanned on a 3T scanner from the TUEF/DZD database. Axial slices of the T1-weighted fast spin echo (FSE), labeled ground-truth and segmentation output of the proposed DCNet are shown. White arrows indicate areas of magnetic field inhomogeneities which were correctly estimated by the DCNet. Quantitative scores over whole cohort are stated in Table 2.

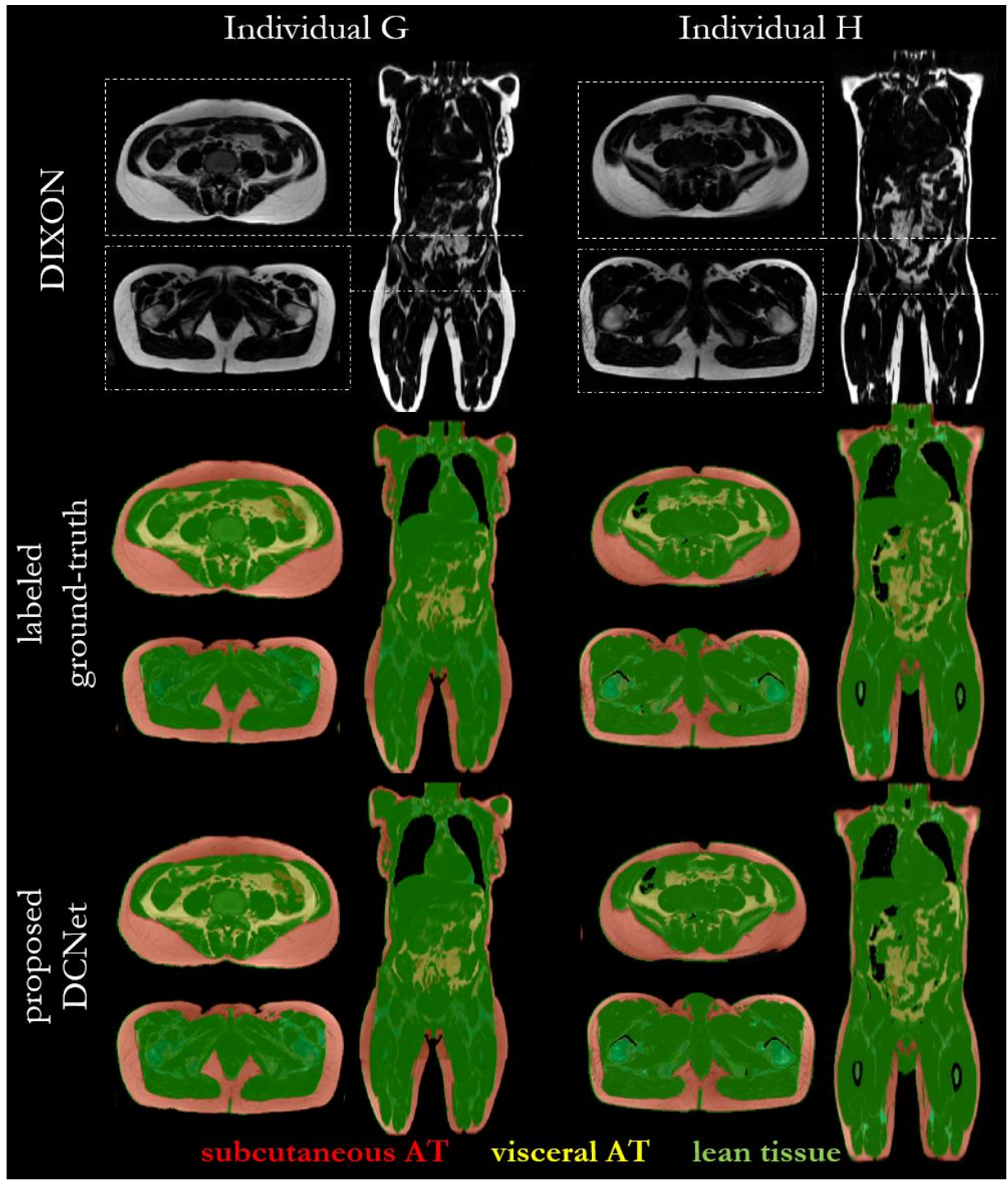

Figure 3. Adipose tissue (AT) segmentation into subcutaneous AT (red), visceral AT (yellow), lean tissue (green) and background (black) in two individuals scanned on a 3T scanner from the NAKO database. Coronal and two exemplary axial slices (at femur head and abdominal) of fat images from the multi-echo chemical-shift (Dixon) sequence, labeled ground-truth and segmentation output of the proposed DCNet are shown. Quantitative scores over whole cohort are stated in Table 2.

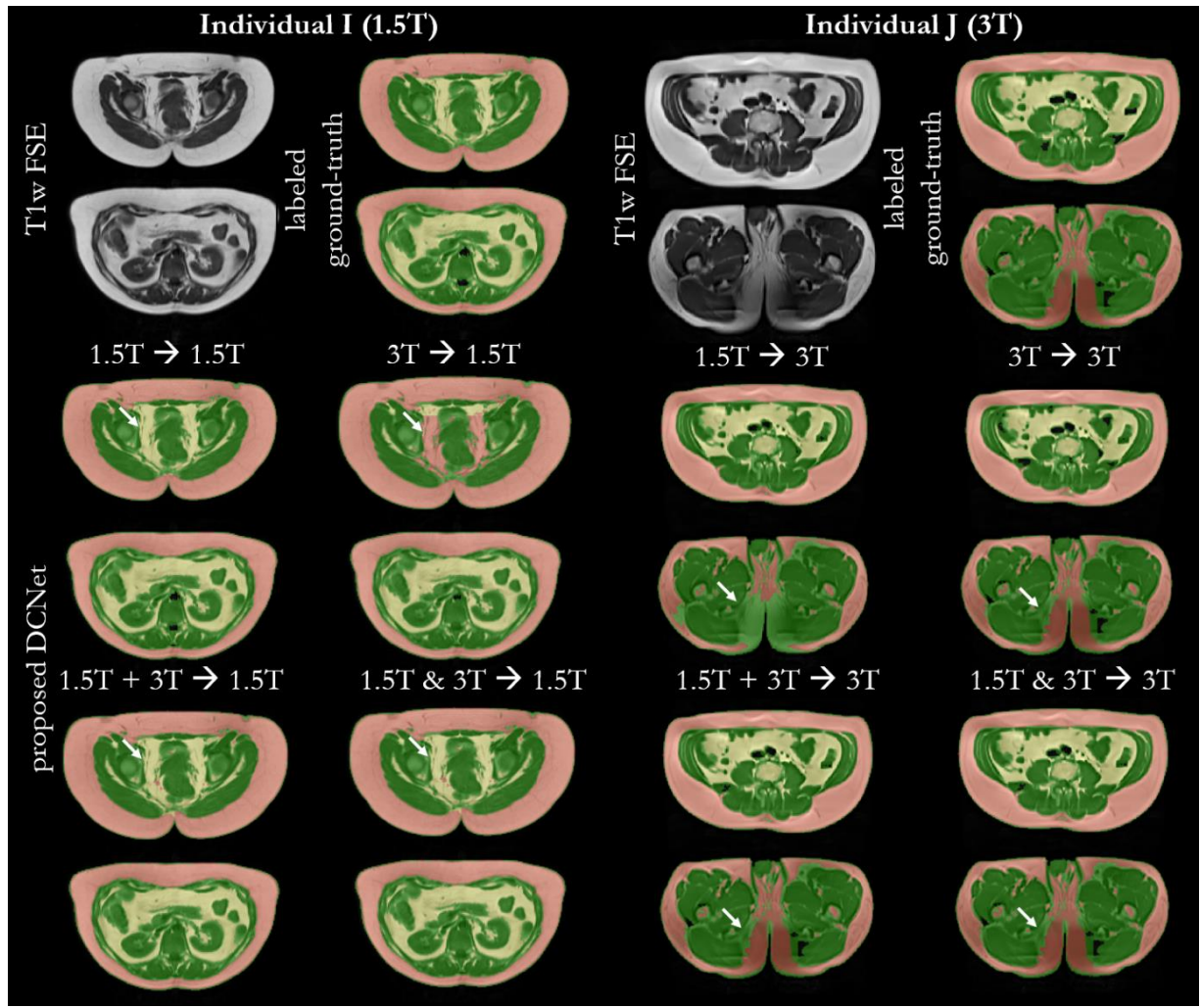

Figure 4. Adipose tissue (AT) segmentation into subcutaneous AT (red), visceral AT (yellow), lean tissue (green) and background (black) in one individual scanned on a 1.5T scanner and in one individual scanned on a 3T scanner from the TUEF/DZD database. Axial slices of the T1-weighted fast spin echo (FSE), labeled ground-truth, and segmentation output of the proposed DCNet are shown. Different training and testing scenarios were investigated to examine intra-database (A→A), inter-database (A→B), transfer learning (A+B → A|B) or multi-database learning (A&B → A|B) for changing imaging scanners and field strengths. The notation A→B|C denotes training on A and testing on B or C. White arrows indicate areas which were falsely classified in some experiments. Quantitative scores over whole cohort are stated in Table 3.

Küstner et al. *Radiology Artificial Intelligence* 2020

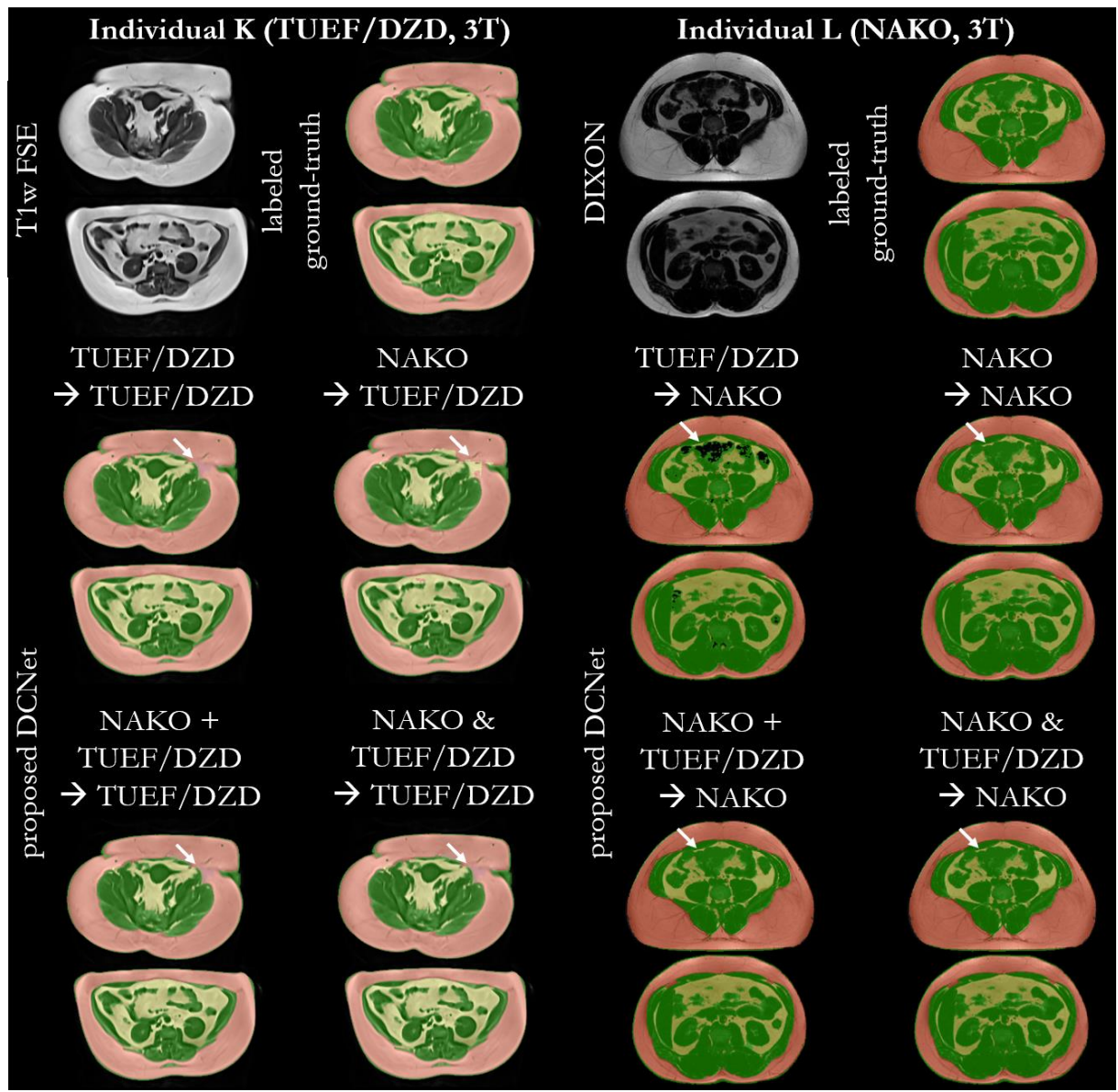

Figure 5. Adipose tissue (AT) segmentation into subcutaneous AT (red), visceral AT (yellow), lean tissue (green) and background (black) in one TUEF/DZD individual scanned on a 3T scanner and in one NAKO individual scanned on a 3T scanner. Axial slices of the T1w fast spin echo (FSE) sequence respectively multi-echo chemical-shift (Dixon) fat images, labeled ground-truth and segmentation output of the proposed DCNet are shown. Different training and testing scenarios were investigated to examine intra-database (A→A), inter-database (A→B), transfer learning (A+B → A|B) or multi-database learning (A&B → A|B) for changing epidemiological cohorts (imaging sequence, resolution, coil arrangements, patient positioning, scanner). The notation A→B|C denotes training on A and testing on B or C. White arrows indicate areas which were falsely classified in some experiments. Quantitative scores over whole cohort are stated in Table 4.

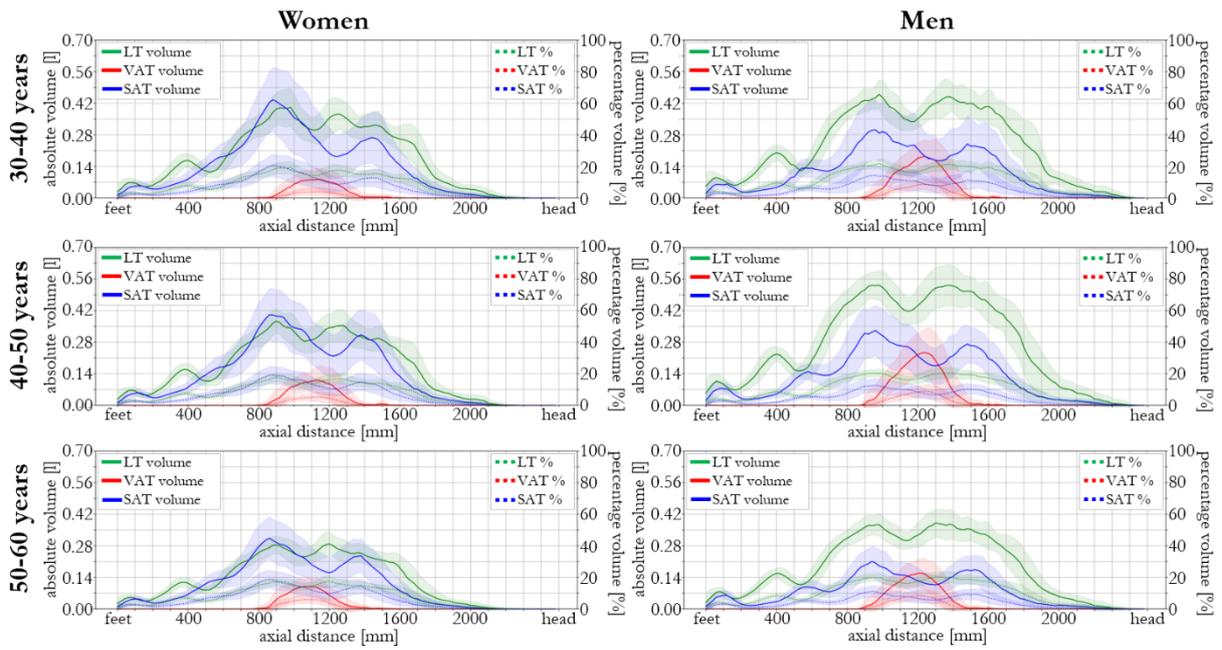

Figure 6. Adipose tissue profiles along head-feet direction over all test cases in TUEF/DZD database grouped according to sex and age. Lean tissue (LT, green), visceral adipose tissue (VAT, red), subcutaneous adipose tissue (SAT, blue) are shown as absolute volume in liters (solid line) and percentage per slice (dotted line). Mean (line) and one standard deviation around mean (colored shaded area) are depicted.

Küstner et al. *Radiology Artificial Intelligence* 2020

**Supplemental Material**

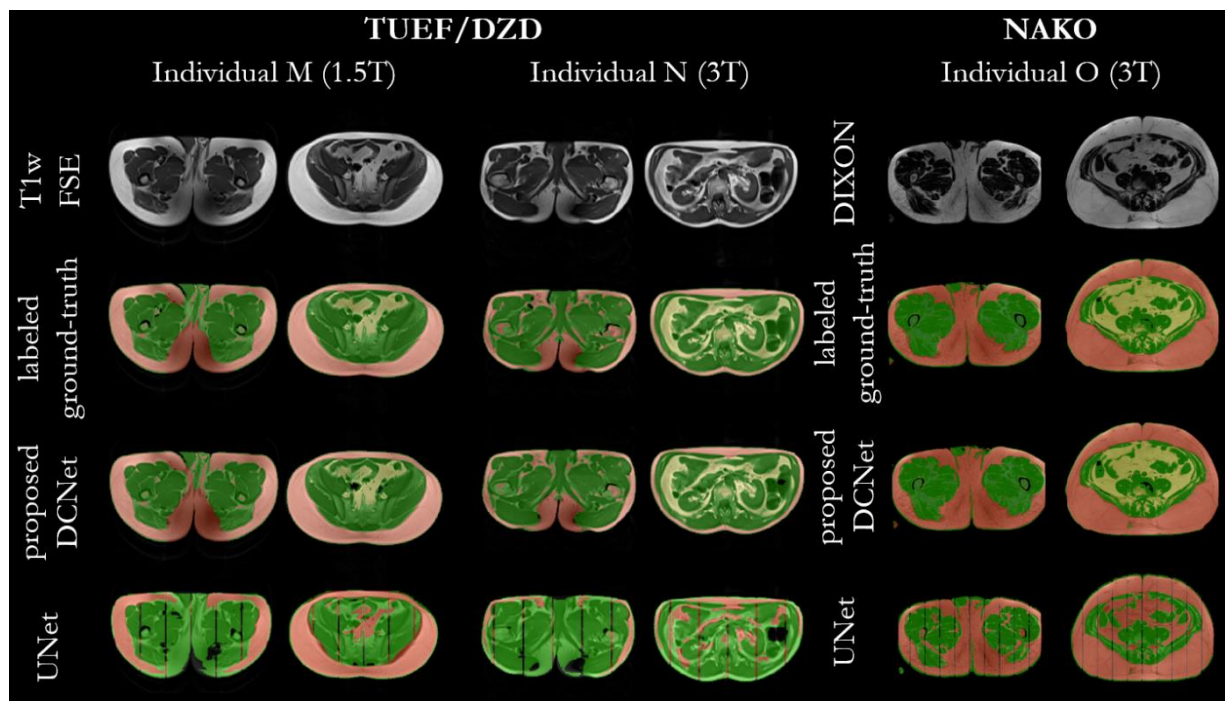

Figure E1: Adipose tissue (AT) segmentation into subcutaneous AT (red), visceral AT (yellow), lean tissue (green) and background (black) in one TUEF/DZD individual scanned on a 1.5T scanner, one TUEF/DZD individual scanned on a 3T scanner and one NAKO individual scanned on a 3T scanner. Axial slices of the T1w fast spin echo (FSE) respectively multi-echo chemical-shift (Dixon) fat images, labeled ground-truth and segmentation output of the proposed DCNet as well as 3D UNet segmentation are shown. Quantitative scores over whole cohort are stated in Table 2.

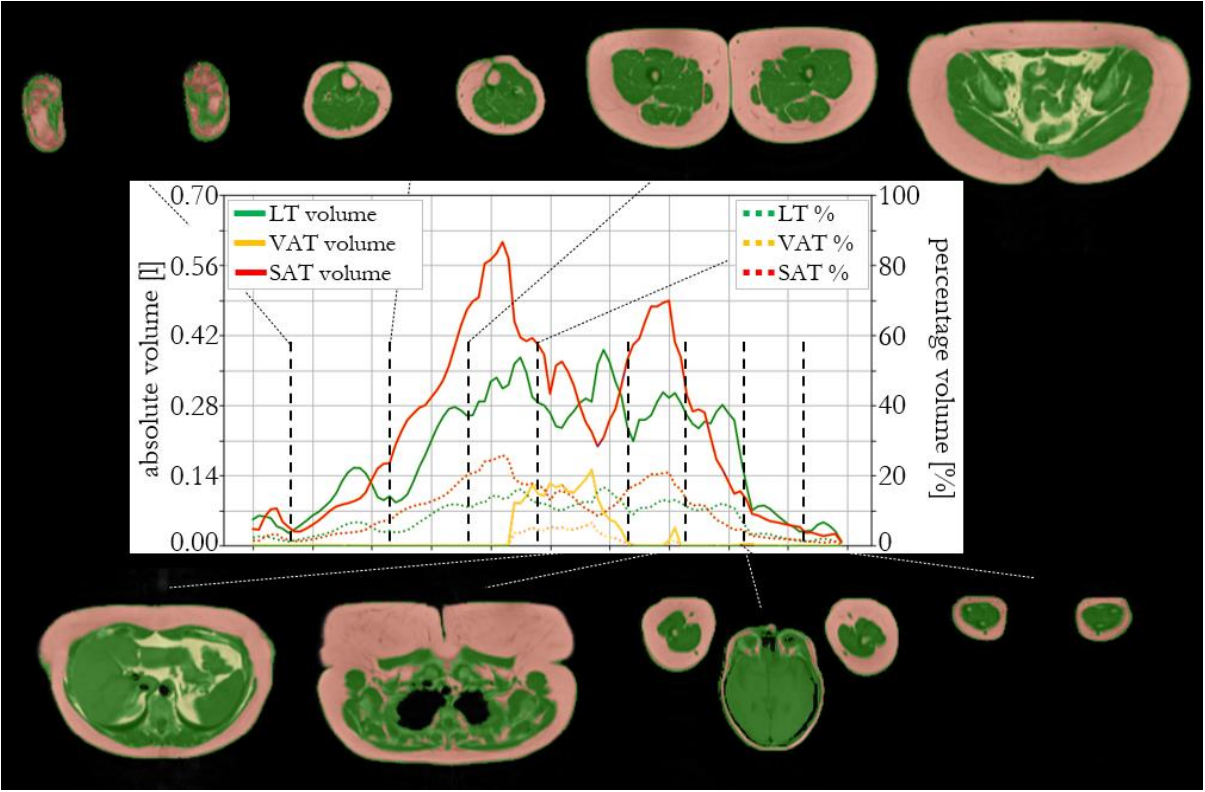

Figure E2: Exemplary adipose tissue head-feet profile of a TUEF/DZD individual with lean tissue (LT: green), visceral adipose tissue (VAT: yellow) and subcutaneous adipose tissue (SAT: red) as absolute volume in liters (sold line) and percentage per slice (dotted line). Corresponding segmented slices of proposed DCNet are shown.

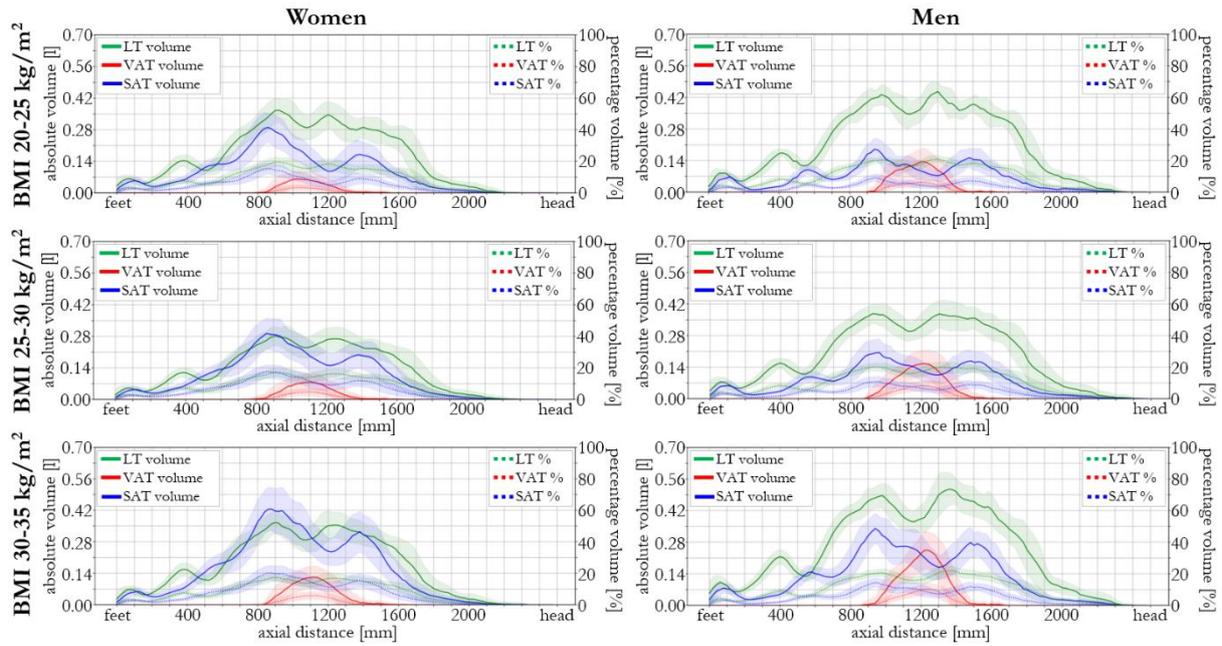

Figure E3: Adipose tissue profiles along head-feet direction over all test individual in TUEF/DZD database grouped according to sex and body mass index (BMI). Lean tissue (LT: green), visceral adipose tissue (VAT: red), subcutaneous adipose tissue (SAT: blue) are shown as absolute volume in liters (sold line) and percentage per slice (dotted line). Mean (line) and one standard deviation around mean (colored shaded area) are depicted.

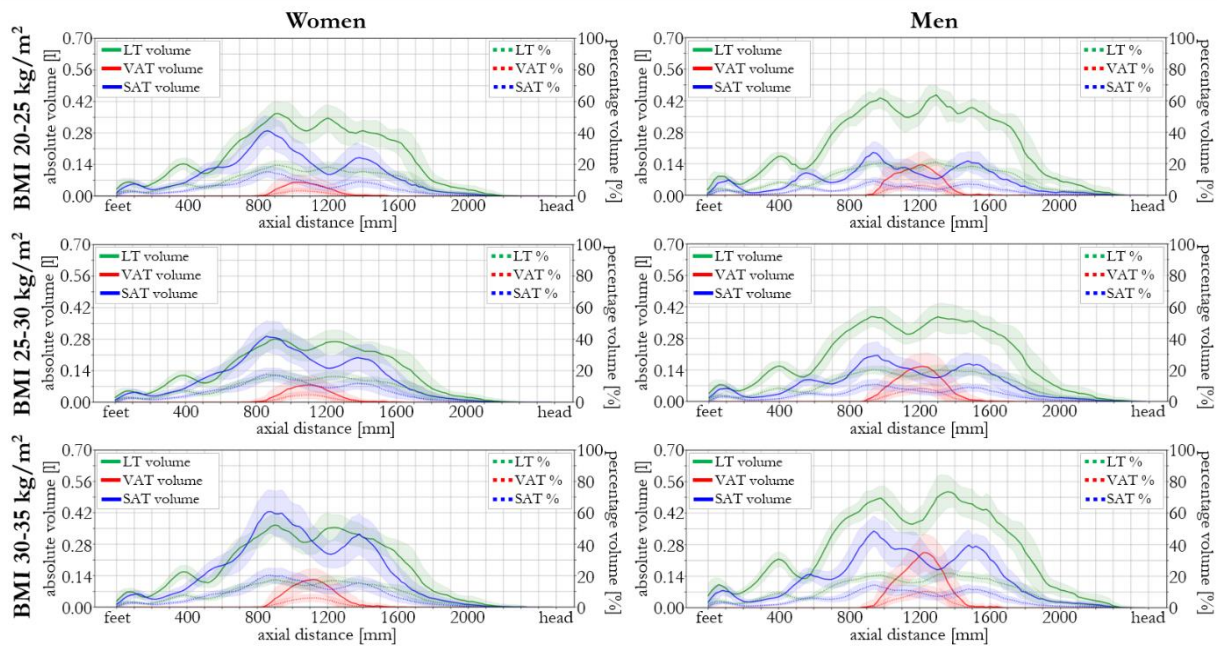

Figure E4: Adipose tissue profiles along head-feet direction over all test individual in NAKO database grouped according to sex and age. Lean tissue (LT: green), visceral adipose tissue (VAT: red), subcutaneous adipose tissue (SAT: blue) are shown as absolute volume in liters (sold line) and percentage per slice (dotted line). Mean (line) and one standard deviation around mean (colored shaded area) are depicted.

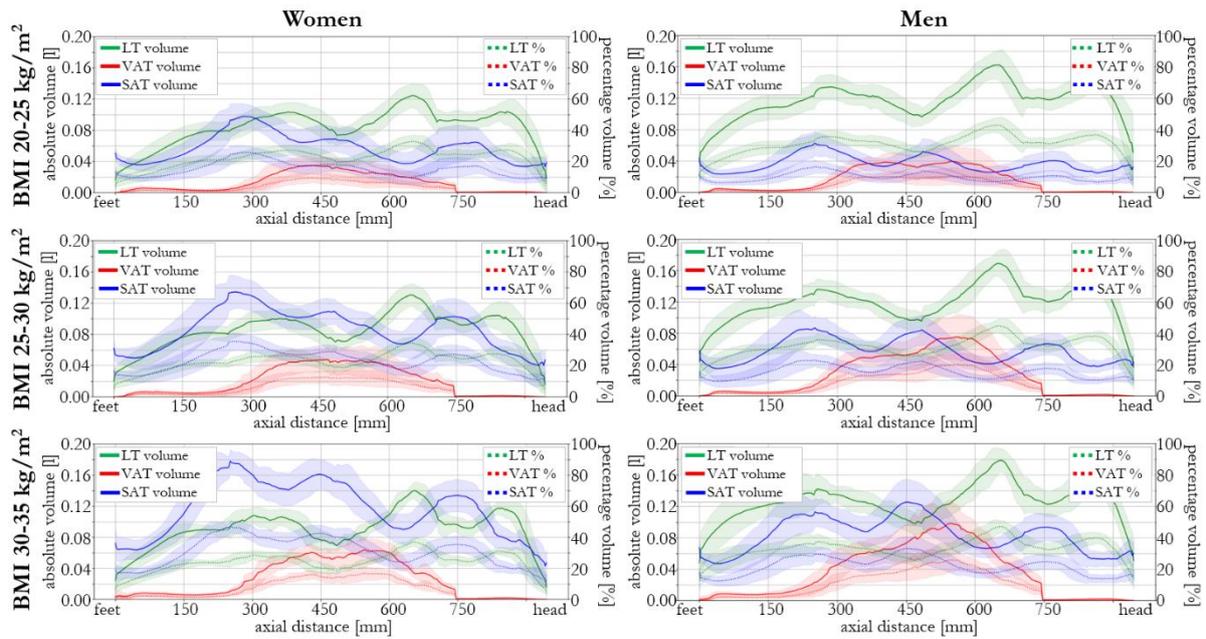

Figure E5: Adipose tissue profiles along head-feet direction over all test individual in NAKO database grouped according to sex and body mass index (BMI). Lean tissue (LT: green), visceral adipose tissue (VAT: red), subcutaneous adipose tissue (SAT: blue) are shown as absolute volume in liters (sold line) and percentage per slice (dotted line). Mean (line) and one standard deviation around mean (colored shaded area) are depicted.

**Appendix E1. Network Architecture**

The general architectural shape is given by UNet and VNet (14, 15) as a backbone for well generalizable segmentation networks. The ascending decoder branch (right-hand side) restores the input shape from the extracted features of the descending encoder branch (left-hand side). Each stage on either side of the branches is composed of a Merge-And-Run block followed by a Transition Layer with down- (encoder) or upsampling (decoder). Each Merge-And-Run block consists of *L* Dense Convolutional nodes in depth which are operated in two parallel pathways which are connected between and within in similar manner than a DenseNet (42). Thus, feature maps are forwarded via residual connections to all subsequent layers where they are merged via summation along channels. This enables the usage of low-level features in deeper layers and provides by that a deep supervision making the network deeper instead of increasing the number of convolutional layers. In combination with Merge-And-Run, an attention-based multi-resolution focusing is obtained that allows to segment small and diverse structures. One Dense Convolutional node is composed of batch normalization (BN), rectified linear unit (ReLU) activation function, $1 \times 1 \times 1$ convolution, BN, ReLU and $3 \times 3 \times 3$ convolution with dyadic varying rates of dilation (1, 2, 4, 8, 1) per dense convolution node to increase the receptive field and to provide a multi-resolution focusing within a stage and over stages (pooling/transposed convolutions). The bottleneck convolution enables a reduction in trainable parameters. The Transition Layers are built up as one Dense Convolutional node followed by a 2 x 2 x 2 max-pooling (TransitionLayerDown) respectively transposed convolution with stride 2 (TransitionLayerUp) to provide downsampling and upsampling of the feature maps. Data is downsampled along the encoder to a size of 1 x 1 x 1 and upsampled back towards its input size in the decoder.

In the deepest stage, a positional encoding is embedded into the learning process in order to provide spatial reference which might have been lost during cropping of the input images and to prevent misclassification at unlikely locations. The absolute 3D position of each cube given in the scanner coordinate system is concatenated to the feature map in the deepest layer.

**Appendix E2. Comparative 3D UNet**

The 3D UNet (14, 15) consists of three encoding and decoding stages and one bottleneck stage with two 3D convolutional layers, batch normalization, and ReLU activation function in each. A kernel size of 3 x 3 x 3 with dyadic feature channel increase (starting at 32) per convolutional layer and a 2 x 2 x 2 max-pooling (encoder) respectively 3D transpose convolution with stride 2 x 2 x 2 (decoder) was performed. The 3D UNet was trained with otherwise same loss function and parameters.

**Appendix E3. Statistical Analysis**

Values indicate true positive (TP), true negative (TN), false positive (FP), and false negative (FN) voxels between predicted segmentation $S$ and ground-truth $G$.

$$\text{Specificity} = \frac{\text{TN}}{\text{TN} + \text{FP}} \quad (1)$$

$$\text{Sensitivity} = \frac{\text{TP}}{\text{TP} + \text{FN}} \quad (2)$$

$$\text{Precision} = \frac{\text{TP}}{\text{TP} + \text{FP}} \quad (3)$$

$$\text{Accuracy} = \frac{\text{TP} + \text{TN}}{\text{TP} + \text{TN} + \text{FP} + \text{FN}} \quad (4)$$

$$\text{Dice} = \frac{2 \cdot |S \cap G|}{|S| + |G|} = \frac{2 \cdot \text{TP}}{2 \cdot \text{TP} + \text{FP} + \text{FN}} \quad (5)$$